\documentclass[letterpaper]{article} 
\usepackage{aaai2026}  
\usepackage{times}  
\usepackage{helvet}  
\usepackage{courier}  
\usepackage[hyphens]{url}  
\usepackage{graphicx} 
\usepackage{natbib}  
\usepackage{caption} 
\usepackage{tikz}
\usetikzlibrary{arrows.meta, positioning, calc, shapes.geometric, shapes.misc, fit, backgrounds, decorations.pathreplacing}
\usepackage{listings}
\usepackage{booktabs}
\usepackage{multirow}
\usepackage{amsmath}
\usepackage{amssymb}
\usepackage{makecell}

\newcommand{\barvaluefont}{\fontsize{7.5}{9}\selectfont}

\newcommand{\Pk}[1]{Pass\texttt{\char94}#1}

\title{Closing the Consistency Gap: Self-Evolving Agents That Learn to Stay on Course}
\author{
    Evelyn Duesterwald\textsuperscript{\rm 1},
    Benjamin Elder\textsuperscript{\rm 1},
    Lilian Ngweta\textsuperscript{\rm 1},
    Shashanka Ubaru\textsuperscript{\rm 2},
    Malgorzata Zimon\textsuperscript{\rm 2}
}
\affiliations{
    \textsuperscript{\rm 1}IBM Software Innovation Lab\\
    \textsuperscript{\rm 2}IBM Research
}

\begin{document}

\maketitle

\begin{abstract}
Large language model (LLM)-powered agents can be accurate \emph{on average} yet unreliable in production, a discrepancy that has been observed but remains largely unaddressed. When given the \emph{same} task five times, a ReAct agent on the AppWorld benchmark using GPT-4.1 succeeds in all five runs only 53\% of the time, even though its per-run pass rate averages 77\%. We call this 24-point shortfall the \emph{consistency gap}, and we argue that addressing it is a precondition for trustworthy AI agent deployment. We present a self-evolving agent framework that reduces this gap by identifying unstable, low-consistency steps in agent trajectories and converting them into episodic memory the agent can draw on in future runs. 
At its core is a \emph{Consistency Analyzer} that pinpoints where and why a trajectory is likely to flip across executions, and a \emph{Guideline Generator} that converts the diagnosis into targeted guidelines, committed to memory and injected into future agent executions on similar tasks.
On AppWorld with ReAct/GPT-4.1, our framework raises the fraction of tasks that succeed in all five runs by \textbf{+16~points} on same-task evaluation and \textbf{+13~points} on similar-task generalization. 
\end{abstract}


\noindent\textbf{Code:} \url{https://github.com/AgentToolkit/altk-evolve}

\section{Introduction}
\label{sec:intro}

Large language model (LLM)-powered agents have become the technology of choice for automating tasks that span from web navigation to API orchestration to interactive coding. Published benchmark accuracy numbers miss an underappreciated caveat: agents can be accurate \emph{on average} but unreliable \emph{on repeat}, since the same task asked of the same agent often produces different outcomes from one run to the next. We call this shortfall the \emph{consistency gap}, and argue that reducing it is a precondition for the trust and reliability required for production deployment.

The gap is often large enough to matter in practice. On the \emph{test\_normal} split of AppWorld~\cite{trivedi2024appworld}, a ReAct~\cite{yao2023react} agent backed by GPT-4.1 achieves a mean pass rate of $77\%$ across five runs (Mean@5), but consistently passes all five runs on only $53\%$ of tasks (\Pk{5}). This $24.4$-point shortfall concentrates in a substantial population of tasks ($\sim$32\% of the benchmark) that mix passes and failures across runs, the scenarios where mean accuracy numbers are least informative about what a user actually experiences.

The design of our framework to address this gap rests on three observations. First, consistency is not the same as accuracy: a high mean pass rate can co-exist with a low consistent pass rate, so treating consistency as a first-class objective requires dedicated metrics (Section~\ref{sec:metrics}), an instrument that estimates it without re-running every task many times (Section~\ref{sec:analyzer}), and a mechanism that targets it specifically (Section~\ref{sec:guidelines}). Second, inconsistency has a measurable token-level fingerprint: at each inference point the model induces a distribution over next tokens, and a \emph{flat} distribution with near-tied tokens lets the agent's decision flip from run to run even under temperature-zero decoding; this can be estimated without model internals by resampling and measuring response variability, the basis of our \emph{Consistency Analyzer}. Third, consistency can be improved by injecting consistency guidance at the prompt level: a short natural-language \emph{consistency guideline} replaces an uncertain, previously flip-prone decision with a stable one the agent makes the same way on every subsequent run - without touching the inference loop, agnostic to agent architecture, and confined to an offline analysis stage.

\begin{figure}[t]
    \centering
    \includegraphics[width=0.47\textwidth]{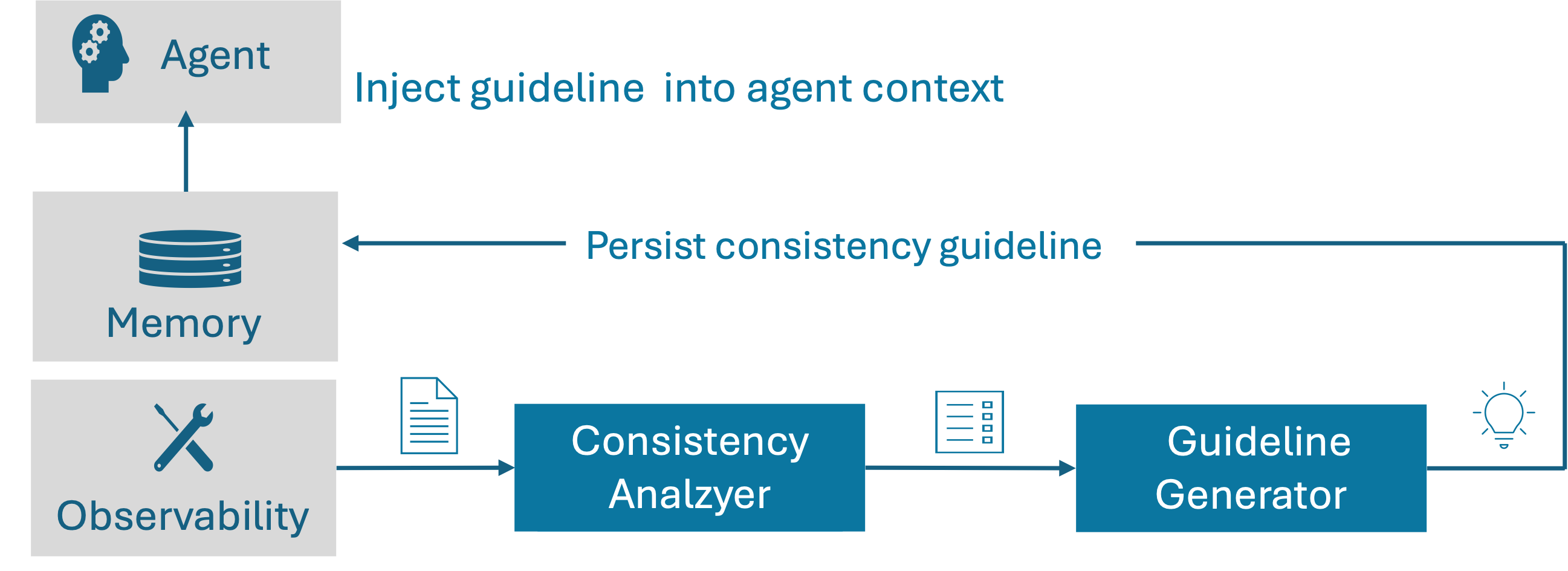}
    \caption{Consistency guideline pipeline: Detect (1) and Generate (2).}
    \label{fig:approach-overview}
\end{figure}

\textbf{Approach.}
Our framework is a two-stage pipeline (Figure~\ref{fig:approach-overview}) that is triggered when a new trajectory becomes available. Both stages run offline. \textbf{Detect}: a black-box Consistency Analyzer resamples each inference point in the trajectory, producing a step- and trajectory-level \emph{consistency scorecard} that pinpoints decisions at risk of flipping, without access to model internals or agent code. \textbf{Generate}: an LLM-based Guideline Generator converts each flip-prone step into one or more candidate \emph{consistency guidelines}, organized (following \citet{fang2026trajectory}) into strategy, recovery, and optimization categories. Generated guidelines are committed to memory, clustered and deduplicated, and retrieved by cosine similarity into the prompt of subsequent similar tasks, stabilizing the inference points that are similar to those previously flagged. We build on the trajectory-informed memory framework of \citet{fang2026trajectory} for storage, clustering, consolidation, and retrieval; the principal cost of this cross-execution learning is a delay before useful guidelines accumulate, bounded by a feedback dynamic in which each guideline shrinks the inconsistent-task population that future guidelines are drawn from.

\textbf{Contributions.}
(1) A formalization and empirical characterization of the consistency gap (Section~\ref{sec:gap}).
(2) A black-box Consistency Analyzer (Section~\ref{sec:analyzer}) that estimates step- and trajectory-level consistency from response variability via configurable resampling; aggregate trajectory consistency also correlates with trajectory pass/fail (AUROC $0.69$), letting it support the generator's own judgment of trajectory outcome where no ground-truth rating is available. (3) Consistency-targeted guideline generation (Section~\ref{sec:guidelines}) 
driven by step-level inconsistency. (4) Empirical results (Section~\ref{sec:evaluation}) on AppWorld with ReAct/GPT-4.1 and GPT-OSS-120B showing $+16$ pp \Pk{5} same-task and $+13$ pp similar-task generalization (168 tasks), plus a sample-budget sensitivity study.

\section{Consistency Gap}
\label{sec:gap}

A common issue with current AI agents is: the same task, asked of the same agent backed by the same model, can produce different outcomes from one run to the next. In this section we make this discrepancy explicit, empirically characterize its magnitude, and sketch its underlying causes.

\subsection{Pass@k, Mean@k, and \Pk{k}}
\label{sec:metrics}

Let $\mathcal{T}$ be a benchmark of agent tasks. For each task $t \in \mathcal{T}$, run the agent $k$ independent times and let $X_t \in \{0, 1\}^k$ record outcome of a run of $t$. Three task-level aggregation metrics of $X_t$ are commonly reported in the agentic literature:

\begin{align}
\mathrm{Pass@}k(t) &\;\triangleq\; \mathbf{1}\!\left[(~\textstyle\sum_{j=1}^{k} X_t^{(j)}) \ge 1\right], \label{eq:passatk}\\[2pt]
\mathrm{Mean@}k(t) &\;\triangleq\; \tfrac{1}{k}\textstyle\sum_{j=1}^{k} X_t^{(j)}, \label{eq:meanatk}\\[2pt]
\text{\Pk{k}}(t) &\;\triangleq\; \mathbf{1}\!\left[(~\textstyle\sum_{j=1}^{k} X_t^{(j)}) = k\right]. \label{eq:passpowk}
\end{align}

\noindent
Pass@$k$ asks whether the agent succeeds \emph{at least once} in $k$ tries, isolating capability from sampling noise. Mean@$k$ is the per-task mean pass rate, the standard reported accuracy. \Pk{k}, by contrast, asks whether the agent succeeds on \emph{all} $k$ tries, isolating \emph{consistency} from capability. For $k$ independent Bernoulli trials with success probability $p$, the three satisfy $\mathbb{E}[\text{\Pk{k}}] = p^k \le \mathbb{E}[\mathrm{Mean@}k] = p \le \mathbb{E}[\mathrm{Pass@}k] = 1-(1-p)^k$, with equality only at $p \in \{0,1\}$. Aggregated over a benchmark, $\mathrm{Pass@}k$, $\mathrm{Mean@}k$ and $\text{\Pk{k}}$ refer to averages of the per-task quantities over $\mathcal{T}$.

\paragraph{Consistency gap.}
We define the \emph{consistency gap} as $\mathrm{Gap@}k(\mathcal{T}) \triangleq \mathrm{Mean@}k(\mathcal{T}) - \text{\Pk{k}}(\mathcal{T})$\footnote{Jha et al.~\cite{jha2026reliability} independently introduce \emph{reliability gap} for the analogous discrepancy between Pass@$k$ and Majority@$k$ in ReAct agents; see Section~\ref{sec:relatedwork}.}, measured in percentage points (pp): it quantifies how much apparent accuracy is unreliable, i.e., may evaporate under repeated deployment. Figure~\ref{fig:consistency-gap} illustrates this for a ReAct agent on AppWorld ($k=5$), plotting Mean@5, \Pk{5}, and the gap between them, per difficulty level and model.

\paragraph{Normalized consistency.}
The raw gap confounds capability with consistency, since it is mechanically bounded above by Mean@$k$: a low-accuracy model can exhibit at most a small absolute gap regardless of how variable its behavior is. We therefore also define \emph{normalized consistency} $\mathrm{Consistency@}k(\mathcal{T}) \triangleq \text{\Pk{k}}(\mathcal{T})/\mathrm{Mean@}k(\mathcal{T}) \in [0,1]$, which rescales \Pk{k} by the model's own mean accuracy: $1$ means the agent passes every run on every task it can ever solve, near $0$ means successes are almost entirely non-reproducible. Substituting gives $\mathrm{Gap@}k = \mathrm{Mean@}k \cdot (1 - \mathrm{Consistency@}k)$, so two agents with identical absolute gaps can have different Consistency@$k$ whenever their mean accuracy differs.

\definecolor{cnsMeanBar}{cmyk}{0.70,0.15,0,0.45}
\definecolor{cnsPassBar}{cmyk}{1.00,0.85,0,0.45}
\definecolor{cnsGapMark}{cmyk}{0,0.65,0.85,0.30}

\newcommand{\gappanel}[2]{%
  \begin{tikzpicture}[
    font=\normalsize,
    bar/.style={draw=none, fill=##1},
    axis/.style={->, >=Stealth, thick},
    ]
    \def\h{2.4}      
    \def\bw{0.32}    
    \def\gw{0.10}    
    \def\groupw{1.6} 
    \def\bottomy{0}

    \draw[axis] (0,0) -- (0,\h+0.25);
    \foreach \pct in {0,20,40,60,80,100} {
      \pgfmathsetmacro{\y}{\pct/100*\h}
      \draw[gray!30] (0,\y) -- (6.8,\y);
      \node[font=\footnotesize, anchor=east] at (-0.05,\y) {\pct\%};
    }
    \node[rotate=90, anchor=south, font=\footnotesize] at (-0.7,\h/2) {pass rate (\%)};

    \def\groups{#1}

    \foreach \i in {0,...,3} {
      \pgfmathsetmacro{\xc}{0.5 + \i*\groupw + \groupw/2}
      \pgfmathsetmacro{\meanv}{\groups[\i][1]}
      \pgfmathsetmacro{\passv}{\groups[\i][2]}
      \pgfmathsetmacro{\meany}{\meanv/100*\h}
      \pgfmathsetmacro{\passy}{\passv/100*\h}

      \pgfmathsetmacro{\xleft}{\xc - \bw - \gw/2}
      \fill[cnsMeanBar] (\xleft,0) rectangle ++(\bw,\meany);
      \node[font=\barvaluefont, anchor=south, fill=white, inner sep=1pt] at (\xleft+\bw/2, \meany) {\pgfmathprintnumber[fixed,fixed zerofill,precision=1]{\meanv}};

      \pgfmathsetmacro{\xleft}{\xc + \gw/2}
      \fill[cnsPassBar] (\xleft,0) rectangle ++(\bw,\passy);

      \pgfmathsetmacro{\passlabeltop}{\passy + 0.12} 
      \node[font=\barvaluefont, anchor=south west, fill=white, inner sep=1pt]
        at (\xleft+\bw+0.05, \passy) {\pgfmathprintnumber[fixed,fixed zerofill,precision=1]{\passv}};

      \pgfmathsetmacro{\gapv}{\meanv - \passv}
      \pgfmathsetmacro{\arrowx}{\xleft + \bw/2}
      \pgfmathsetmacro{\arrowloy}{\passlabeltop}
      \pgfmathsetmacro{\arrowlen}{\meany - \arrowloy}
      \pgfmathsetmacro{\gapymid}{(\meany+\arrowloy)/2}
      \pgfmathtruncatemacro{\drawarrow}{(\arrowlen > 0.20) ? 1 : 0}
      \ifnum\drawarrow=1
        \draw[<->, cnsGapMark, semithick, >={Stealth[length=3pt,width=2pt]}]
          (\arrowx, \arrowloy) -- (\arrowx, \meany);
      \fi
      \node[font=\barvaluefont, text=cnsGapMark, anchor=west, fill=white, inner sep=1pt]
        at (\arrowx + 0.05, \gapymid) {\pgfmathprintnumber[fixed,fixed zerofill,precision=1]{\gapv}};

    }

    \node[font=\footnotesize, align=center] at (1.3, -0.50) {Aggregate\\(168)};
    \node[font=\footnotesize, align=center] at (2.9, -0.50) {Easy\\(57)};
    \node[font=\footnotesize, align=center] at (4.5, -0.50) {Medium\\(48)};
    \node[font=\footnotesize, align=center] at (6.1, -0.50) {Hard\\(63)};

    \fill[cnsMeanBar] (1.0, \h+0.45) rectangle ++(0.3,0.2);
    \node[font=\footnotesize, anchor=west] at (1.35, \h+0.55) {Mean@5};
    \fill[cnsPassBar] (3.0, \h+0.45) rectangle ++(0.3,0.2);
    \node[font=\footnotesize, anchor=west] at (3.35, \h+0.55) {\Pk{5}};
    \draw[<->, cnsGapMark, semithick, >={Stealth[length=3pt,width=2pt]}] (4.7, \h+0.40) -- (4.7, \h+0.70);
    \node[font=\footnotesize, text=cnsGapMark, anchor=west] at (4.85, \h+0.55) {gap (pp)};

    \node[font=\normalsize, anchor=north] at (3.4, -1.00) {#2};
  \end{tikzpicture}%
}

\begin{figure*}[t]
\centering
\gappanel{{
  {"Aggregate (168)", 77.4, 53.0},
  {"Easy (57)", 94.7, 77.2},
  {"Medium (48)", 77.1, 52.1},
  {"Hard (63)", 61.9, 31.7}
}}{(a) ReAct / GPT-4.1}\hspace{0.5cm}%
\gappanel{{
  {"Aggregate (168)", 33.9, 10.1},
  {"Easy (57)", 66.7, 28.1},
  {"Medium (48)", 27.1, 2.1},
  {"Hard (63)", 9.5, 0.0}
}}{(b) ReAct / GPT-OSS-120B}
\caption{The \emph{consistency gap} on AppWorld for a ReAct agent backed by (a) GPT-4.1 and (b) GPT-OSS-120B.}
\label{fig:consistency-gap}
\end{figure*}

\subsection{Empirical Characterization of the Consistency Gap}
\label{sec:gap-evidence}

We measure the consistency metrics for a ReAct~\cite{yao2023react} agent backed by two LLMs of different capabilities, GPT-4.1 and GPT-OSS-120B, evaluated on the 168-task \texttt{test\_normal} split of the AppWorld benchmark~\cite{trivedi2024appworld} with $k = 5$ runs per task and the standard AppWorld grader (Figure~\ref{fig:consistency-gap}). 

\paragraph{Absolute gap.}
The absolute gap is large for both models ($24.4$\,pp for GPT-4.1 and $23.8$\,pp for GPT-OSS-120B), despite the wide capability difference but distributes differently across difficulty. With GPT-4.1 it grows monotonically ($17.5$/$25.0$/$30.2$\,pp for easy/medium/hard): harder tasks are both less likely to succeed and disproportionately likely to flip. With GPT-OSS-120B the pattern inverts, peaking on easy ($38.6$\,pp) and shrinking on hard ($9.5$\,pp) because the hard-task Mean@5 itself collapses to $9.5\%$, mechanically capping the gap. In both cases the consistent fraction of the benchmark falls far below mean accuracy, $53\%$ vs.\ Mean@5 of $77\%$ for GPT-4.1, only $10\%$ vs.\ $34\%$ for GPT-OSS-120B, so standard accuracy numbers are substantially uninformative about what a deployed-agent user actually experiences.

\paragraph{Normalized consistency.}
Normalizing by each model's own mean accuracy removes the capability ceiling. GPT-4.1's aggregate consistency is $0.69$, declining to $0.51$ on hard tasks: even where it can occasionally solve a task, it reliably delivers all-five-pass only about half the time.  
GPT-OSS-120B is more severe: aggregate consistency is only $0.30$, collapsing to $0.00$ on hard tasks - it \emph{never} passes all five runs on a hard task it can occasionally solve. The $2\times$ difference ($0.69$ vs.\ $0.30$) shows GPT-OSS-120B's successes are almost entirely non-reproducible, while GPT-4.1 retains meaningful consistency across difficulty levels.

The gap is unlikely to be idiosyncratic to a single agent or model: in informal testing we have also observed similar patterns with other agent architectures (e.g., CUGA~\cite{marreed2025cuga}) and other models, though we have not quantified the magnitude of the gap in these settings.

These sources of inconsistency trace back to the shape of the model's token-probability distribution at each inference point (Section~\ref{sec:analyzer}), motivating our approach: identify the inference points with flat response distributions and generate prompt-level guidance to sharpen them in future executions.

\section{Consistency Guideline Generation}
\label{sec:approach}

We build on the trajectory-informed agentic memory framework of \citet{fang2026trajectory}\footnote{The Detect-and-Generate pipeline has been contributed as a feature of the open-source \textsc{altk-evolve} repository~\cite{altkevolve2026}.}, 
which treats episodic memory as a primary mechanism for agent self-improvement: an LLM extracts structured \emph{tips} (guidelines) from a raw trajectory in three categories - \emph{strategy}, \emph{recovery}, \emph{optimization} - clusters them into a dual-indexed memory (embedding plus category/priority/context/provenance metadata), and retrieves them at runtime by cosine-similarity or LLM-guided query. We inherit this machinery unchanged; our contribution is a new upstream stage: the pipeline of Figure~\ref{fig:approach-overview} that decides \emph{what} guidelines to extract, targeted at consistency rather than outcome alone.

The approach operates as two coupled loops. The \emph{operational loop} executes user tasks: the agent runs, optionally augmented with guidelines retrieved from memory, and an observability layer captures the full trajectory. The \emph{offline analysis loop} transforms each captured trajectory into new guidelines via the Detect and Generate stages of Figure~\ref{fig:approach-overview}, which enter this dual-indexed episodic memory and are retrieved as described above. Unlike the prior framework, generation here is grounded in the consistency scorecard rather than solely left to the generating LLM's judgment: a step that succeeded but exhibited high response variability is just as eligible for guideline generation as one that observably failed. The two loops form a self-evolving cycle: as trajectory experience accumulates, the analyzer surfaces increasingly fine-grained sources of inconsistency and subsequent runs become measurably more consistent. 

\subsection{Consistency Analyzer}
\label{sec:analyzer}

The Consistency Analyzer takes a single recorded agent trajectory $T = \langle s_1, s_2, \ldots, s_n \rangle$ and produces a \emph{consistency scorecard}: a per-step consistency score $C(s_i) \in [0, 1]$ together with an aggregate trajectory score $C(T) \in [0, 1]$, plus diagnostic metadata identifying which steps are most likely to flip on a future run. The analyzer is black-box in the sense that it only re-invokes the underlying LLM with the agent's recorded prompts, with no access to model logits, internal activations, or agent source code, making it applicable to any agent built on a hosted LLM endpoint.

\subsubsection{Why Greedy Decoding Is Not Enough}
\label{sec:why-resample}

Inconsistency can arise at LLM inference points and is governed by the shape of the token-probability distribution there. Sharp distributions place most mass on a single token and are \emph{resilient} to noise: minor perturbations from GPU floating-point arithmetic, request batching, or tokenizer state are not enough to reorder the top token. Flat distributions, by contrast, place comparable mass on several near-tied tokens and are \emph{vulnerable} to that same noise: a small perturbation is enough to change which token comes out on top. This is why the problem cannot be fixed by decoding strategy alone: switching to greedy decoding, or to sampling with a fixed seed, only fixes how a given probability distribution is turned into a token choice, it does nothing about the fact that, on a hosted endpoint, the probabilities themselves shift slightly from run to run due to platform noise. So the same prompt to the same model can still yield different token choices on different runs, even under temperature-zero (greedy) decoding.
Critically, the confidence of a single response (e.g., the per-token log-probability of the chosen path) reveals only how confident the model was in the path it took, not how close the runners-up were, that is, how resilient the decision actually is to this noise. Predicting whether a step will flip on a future run therefore requires estimating the \emph{shape} of the distribution over agent-meaningful outcomes, not just the depth of the one path taken, an estimate we obtain via resampling.

\subsubsection{Step Consistency via Resampling}
\label{sec:step-consistency}

For each inference step $s_i$, the analyzer reissues the recorded prompt to the same LLM endpoint $N$ times (typically temperature $\le 0.5$), yielding a response set $D_i = \{r_i^{(1)}, \ldots, r_i^{(N)}\}$ that constitutes the empirical estimate of the response distribution. The step consistency score $C(s_i) \in [0,1]$ measures how concentrated $D_i$ is around its mode, using one of three primitive measures or a structured combination, selected by the detected response type.

\paragraph{Free-text and code responses.}
For free natural-language text (e.g., a reasoning trace) or program code, we estimate step consistency as mean pairwise cosine similarity of SBERT~\cite{reimers2019sbert} sentence (or code) embeddings:
\begin{equation}
C(s_i) \;=\; \frac{2}{N(N{-}1)} \sum_{j < k} \cos~\!\bigl(\mathbf{e}(r_i^{(j)}),\, \mathbf{e}(r_i^{(k)})\bigr),
\label{eq:freetext-consistency}
\end{equation}
where $\mathbf{e}(r)$ is the (code) embedding of response $r$. Embedding-space similarity rewards semantic equivalence despite differing syntax, so paraphrases and functionally-equivalent code are correctly treated as consistent.

\paragraph{Categorical responses.}
For categorical values such as a tool name, enum argument, or structured identifier, step consistency is the mean pairwise Jaccard similarity over tokenized values, 
\begin{equation}
C(s_i) = \frac{2}{N(N-1)} \sum_{j<k} J(r_i^{(j)}, r_i^{(k)}), 
\end{equation}
where $J(a,b) = |V(a) \cap V(b)|\,/\,|V(a) \cup V(b)|$ over token sets $V(\cdot)$. Unlike embedding-based measures, Jaccard penalizes any token-level mismatch, appropriate where exact match is semantically required: a different argument key or enum value can produce a different downstream outcome.

\paragraph{Structured combination responses.}
Semi-structured responses (e.g., a JSON object with a free-text \texttt{thoughts} field alongside categorical \texttt{action}/\texttt{action\_input} fields) are decomposed into fields $\mathcal{F}_i$, each mapped to one of the primitive types above, and combined as a weighted sum:
\begin{equation} 
C(s_i) = \sum_{f \in \mathcal{F}_i} \delta_f \cdot \mathrm{sim}_f(D_i^{f}), \sum_f \delta_f = 1, 
\end{equation}
where $D_i^{f}$ projects the $N$ samples onto field $f$, $\delta_f$ is a configurable importance weight (zero to suppress irrelevant fields), and $\mathrm{sim}_f$ is the primitive measure for that field's type. This lets the analyzer tolerate paraphrastic and code-level noise while still flagging meaningful divergence in fields whose exact value determines the agent's next action.

\subsubsection{Trajectory Consistency Aggregation}
\label{sec:traj-aggregation}

Step-level scores alone do not predict whether the trajectory as a whole is at risk, so we aggregate them into a trajectory-level score by mean aggregation: $C_{\mathrm{mean}}(T) = \frac{1}{n} \sum_{i=1}^{n} C(s_i)$. Mean consistency is fast to compute and, as we show in Section~\ref{sec:evaluation}, provides a useful predictor of trajectory outcome. Its limitation is that it treats steps as independent: a high-consistency step that nonetheless propagates an upstream low-consistency decision is not penalized for its dependence on that decision.

\subsection{Guideline Generation}
\label{sec:guidelines}

The Consistency Analyzer surfaces \emph{where} an agent is at risk of flipping; the Guideline Generator turns that diagnosis into a short, agent-readable instruction targeted at reducing the uncertainty that caused it in the first place. To integrate with the memory framework of \citet{fang2026trajectory}, guidelines follow the same strategy/recovery/optimization taxonomy, adapted for consistency: a \emph{strategy} guideline enforces a pattern that was correct despite high variability (the agent guessed right); a \emph{recovery} guideline steers around a step whose variability correlates with failure; and an \emph{optimization} guideline instructs the agent to skip or simplify a highly variable step that is not strictly necessary for task completion.


Given a trajectory $T$ and its consistency scorecard, a step is flagged as inconsistent if its step consistency $C(s_i)$ falls below a threshold $\theta_C$ (we use $\theta_C = 0.85$) and it lies on the agent's decision path. The trajectory, with its flagged steps marked, is passed to an LLM-based generator, which is instructed to generate guidelines that specifically target the flagged inconsistent steps. Appendix~\ref{app:guideline-examples} shows example guidelines generated using GPT-4.1 from the trajectory and scorecard of an AppWorld task.


Generated guidelines enter the dual-indexed episodic memory with metadata for category, priority, application context, task category and provenance. At runtime, guidelines are retrieved by either cosine similarity against the incoming task description or by an LLM-guided selector that constructs a metadata-filtered query from the task. 
We refer to \citet{fang2026trajectory} for the full details of storage management and retrieval, including deduplication, semantic clustering, and LLM-based merging of overlapping guidelines. 
\section{Evaluation}
\label{sec:evaluation}

We evaluate our framework along three dimensions relevant to production deployment: whether consistency guidelines improve repeat-run reliability on the \emph{same} task; whether they generalize to \emph{similar} tasks without per-task tuning, the common case once a system has accumulated experience across many related tasks; and whether the analyzer's aggregate trajectory consistency is a reliable enough proxy for trajectory outcome to support the guideline generator's own judgment when no ground-truth rating is available.

\subsection{Experimental Setup}
\label{sec:eval-setup}

We evaluate on AppWorld~\cite{trivedi2024appworld}, an interactive coding-agent benchmark in which an agent operates a controlled simulation of personal apps via Python tool calls, graded by the AppWorld harness, on the \emph{test\_normal} split (168 tasks).
AppWorld groups tasks into \emph{scenarios} of three variants each, which we exploit below to test generalization to similar tasks. We evaluate a ReAct~\cite{yao2023react} agent backed by GPT-4.1 and GPT-OSS-120B, running each task five times and reporting \Pk{5}, Mean@5, and (where relevant) Pass@5 as defined in Section~\ref{sec:metrics}. 
All experiments were run against cloud-hosted model endpoints: GPT-4.1 on Microsoft Azure and GPT-OSS-120B on Amazon Web Services (AWS).

The consistency analyzer parses the ReAct agent response at each step to extract actions as categorical pairs: (\texttt{API\_name}, \texttt{API\_params}). Each extracted pair is scored by average pairwise Jaccard similarity, with a default resampling budget of $N=30$ at temperature $0.5$. For the same-task and similar-task evaluations, guidelines are generated from a single baseline trajectory per task by prompting the same model that produced it (acting as judge on its own scorecard), and retrieved at runtime by cosine similarity.

\subsection{Same-Task Consistency Improvement}
\label{sec:eval-same-task}

We first evaluate whether guidelines generated from one baseline trajectory of a task improve consistency on repeat runs of that same task; results for these experiments are illustrated in Figure~\ref{fig:closing-gap}.

\paragraph{GPT-4.1.}

Consistency guidelines lift aggregate \Pk{5} by $+16.0$ pp ($53.0\% \to 69.0\%$, a $30\%$ relative gain), largest in absolute terms on Medium tasks ($+22.9$ pp) and in relative terms on Hard tasks ($+14.3$ pp on a $31.7\%$ baseline, a $45\%$ relative lift). Mean@5 is never degraded, rising by $+3.6$ pp aggregate ($+1.6$ to $+6.2$ pp per difficulty level), supporting the design hypothesis that consistency-targeted memory improves $\text{\Pk{k}}$ \emph{without} sacrificing $\mathrm{Mean@}k$. Normalized consistency rises by $+0.17$ aggregate, most on Medium ($+0.22$) and Hard ($+0.21$), where the baseline was most volatile. Full per-difficulty numbers, including similar-task columns, are in Table~\ref{tab:results-gpt41} (Appendix~\ref{app:results}).

\paragraph{GPT-OSS-120B.}

Starting from a substantially lower baseline (\Pk{5}$=10.1\%$, Mean@5$=33.9\%$), consistency guidelines improve aggregate \Pk{5} by $+6.0$ pp and Mean@5 by $+4.8$ pp. The relative \Pk{5} gain ($59\%$) is nearly twice that of GPT-4.1 ($30\%$), indicating the framework extracts more of the available headroom when the baseline is lower. Gains concentrate where the model already has meaningful pass rates: Easy improves by $+8.7$ pp \Pk{5}, while Hard, starting from $0\%$, gains only $+1.6$ pp. Normalized consistency rises from $0.30$ to $0.42$ aggregate, most on Medium ($+0.28$). Full per-difficulty numbers are in Table~\ref{tab:results-oss} (Appendix~\ref{app:results}).

\paragraph{Consistency Gap.}
Figure~\ref{fig:closing-gap} visualizes gap-reduction under same-task guidelines versus baseline. For GPT-4.1 (a), the aggregate gap to perfect consistency narrows from $0.32$ to $0.15$ ($53\%$ reduction) across every difficulty level. For GPT-OSS-120B (b), baseline gaps are far larger ($0.70$ aggregate, $1.00$ on Hard) and guidelines make meaningful but partial inroads (to $0.58$ aggregate): the model's low raw capability caps how much guidelines alone can recover.

\definecolor{cnsBaseGray}{cmyk}{0,0,0,0.55}
\definecolor{cnsTealMed}{cmyk}{0.60,0,0.20,0.50}
\definecolor{cnsTealDark}{cmyk}{0.85,0,0.35,0.55}
\definecolor{cnsGapPlum}{cmyk}{0.35,0.75,0,0.30}

\begin{figure*}[t]
\centering

\begin{tikzpicture}[
  font=\normalsize,
  axis/.style={->, >=Stealth, thick},
  ]
  \def\h{2.1}   

  \draw[axis] (0,0) -- (0,\h+0.25);

  \foreach \val/\lbl in {0/0.0, 0.2/0.2, 0.4/0.4, 0.6/0.6, 0.8/0.8} {
    \pgfmathsetmacro{\y}{\val*\h}
    \draw[gray!30] (0,\y) -- (6.2,\y);
    \node[font=\footnotesize, anchor=east] at (-0.05,\y) {\lbl};
  }

  \draw[dashed, gray!60, thick] (0,\h) -- (6.2,\h);
  \node[font=\footnotesize, anchor=east] at (-0.05,\h) {1.0};

  \node[rotate=90, anchor=south, font=\footnotesize] at (-0.75,\h/2)
    {consistency (\Pk{5}/Mean@5)};

  \foreach \i/\vbase/\vsame/\vsim in {
    0/0.68/0.85/0.83,
    1/0.82/0.91/0.93,
    2/0.68/0.90/0.80,
    3/0.51/0.73/0.71%
  } {
    \pgfmathsetmacro{\xc}{0.8 + \i*1.5}
    \pgfmathsetmacro{\ybase}{\vbase*\h}
    \pgfmathsetmacro{\ysame}{\vsame*\h}
    \pgfmathsetmacro{\ysim}{\vsim*\h}

    \fill[cnsBaseGray]  (\xc-0.435, 0) rectangle ++(0.25, \ybase);
    \fill[cnsTealMed]   (\xc-0.125, 0) rectangle ++(0.25, \ysame);
    \fill[cnsTealDark]  (\xc+0.185, 0) rectangle ++(0.25, \ysim);

    \pgfmathsetmacro{\xarrow}{\xc - 0.310}
    \pgfmathsetmacro{\gapval}{1.0 - \vbase}
    \pgfmathsetmacro{\gapymid}{(\ybase + \h) / 2}
    \draw[<->, cnsGapPlum, semithick, >={Stealth[length=3pt,width=2pt]}]
      (\xarrow, \ybase+0.07) -- (\xarrow, \h-0.07);
    \node[font=\footnotesize, text=cnsGapPlum, anchor=east, fill=white, inner sep=0.5pt]
      at (\xarrow - 0.04, \gapymid)
      {\pgfmathprintnumber[fixed,fixed zerofill,precision=2]{\gapval}};
  }

  \node[font=\footnotesize, align=center] at (0.8, -0.50) {Aggregate\\(168)};
  \node[font=\footnotesize, align=center] at (2.3, -0.50) {Easy\\(57)};
  \node[font=\footnotesize, align=center] at (3.8, -0.50) {Medium\\(48)};
  \node[font=\footnotesize, align=center] at (5.3, -0.50) {Hard\\(63)};

  \fill[cnsBaseGray]   (0.50, \h+0.42) rectangle ++(0.22,0.18);
  \node[font=\footnotesize, anchor=west] at (0.76, \h+0.51) {Baseline};
  \fill[cnsTealMed]    (2.30, \h+0.42) rectangle ++(0.22,0.18);
  \node[font=\footnotesize, anchor=west] at (2.56, \h+0.51) {Same Task};
  \fill[cnsTealDark]   (4.10, \h+0.42) rectangle ++(0.22,0.18);
  \node[font=\footnotesize, anchor=west] at (4.36, \h+0.51) {Similar Task};
  \draw[<->, cnsGapPlum, semithick, >={Stealth[length=3pt,width=2pt]}] (5.95, \h+0.37) -- (5.95, \h+0.67);
  \node[font=\footnotesize, text=cnsGapPlum, anchor=west] at (6.09, \h+0.51) {gap};

  \node[font=\normalsize, anchor=north] at (3.0, -1.00) {(a) ReAct / GPT-4.1};

\end{tikzpicture}%
\hspace{0.7cm}%
\begin{tikzpicture}[
  font=\normalsize,
  axis/.style={->, >=Stealth, thick},
  ]
  \def\h{2.1}

  \draw[axis] (0,0) -- (0,\h+0.25);

  \foreach \val/\lbl in {0/0.0, 0.2/0.2, 0.4/0.4, 0.6/0.6, 0.8/0.8} {
    \pgfmathsetmacro{\y}{\val*\h}
    \draw[gray!30] (0,\y) -- (6.2,\y);
    \node[font=\footnotesize, anchor=east] at (-0.05,\y) {\lbl};
  }

  \draw[dashed, gray!60, thick] (0,\h) -- (6.2,\h);
  \node[font=\footnotesize, anchor=east] at (-0.05,\h) {1.0};

  \node[rotate=90, anchor=south, font=\footnotesize] at (-0.75,\h/2)
    {consistency (\Pk{5}/Mean@5)};

  \foreach \i/\vbase/\vsame/\vsim in {
    0/0.30/0.42/0.46,
    1/0.42/0.49/0.58,
    2/0.08/0.36/0.27,
    3/0.00/0.13/0.29%
  } {
    \pgfmathsetmacro{\xc}{0.8 + \i*1.5}
    \pgfmathsetmacro{\ybase}{\vbase*\h}
    \pgfmathsetmacro{\ysame}{\vsame*\h}
    \pgfmathsetmacro{\ysim}{\vsim*\h}

    \fill[cnsBaseGray]  (\xc-0.435, 0) rectangle ++(0.25, \ybase);
    \fill[cnsTealMed]   (\xc-0.125, 0) rectangle ++(0.25, \ysame);
    \fill[cnsTealDark]  (\xc+0.185, 0) rectangle ++(0.25, \ysim);

    \pgfmathsetmacro{\xarrow}{\xc - 0.310}
    \pgfmathsetmacro{\gapval}{1.0 - \vbase}
    \pgfmathsetmacro{\gapymid}{(\ybase + \h) / 2}
    \draw[<->, cnsGapPlum, semithick, >={Stealth[length=3pt,width=2pt]}]
      (\xarrow, \ybase+0.07) -- (\xarrow, \h-0.07);
    \node[font=\footnotesize, text=cnsGapPlum, anchor=east, fill=white, inner sep=0.5pt]
      at (\xarrow - 0.04, \gapymid)
      {\pgfmathprintnumber[fixed,fixed zerofill,precision=2]{\gapval}};
  }

  \node[font=\footnotesize, align=center] at (0.8, -0.50) {Aggregate\\(168)};
  \node[font=\footnotesize, align=center] at (2.3, -0.50) {Easy\\(57)};
  \node[font=\footnotesize, align=center] at (3.8, -0.50) {Medium\\(48)};
  \node[font=\footnotesize, align=center] at (5.3, -0.50) {Hard\\(63)};

  \fill[cnsBaseGray]   (0.50, \h+0.42) rectangle ++(0.22,0.18);
  \node[font=\footnotesize, anchor=west] at (0.76, \h+0.51) {Baseline};
  \fill[cnsTealMed]    (2.30, \h+0.42) rectangle ++(0.22,0.18);
  \node[font=\footnotesize, anchor=west] at (2.56, \h+0.51) {Same Task};
  \fill[cnsTealDark]   (4.10, \h+0.42) rectangle ++(0.22,0.18);
  \node[font=\footnotesize, anchor=west] at (4.36, \h+0.51) {Similar Task};
  \draw[<->, cnsGapPlum, semithick, >={Stealth[length=3pt,width=2pt]}] (5.95, \h+0.37) -- (5.95, \h+0.67);
  \node[font=\footnotesize, text=cnsGapPlum, anchor=west] at (6.09, \h+0.51) {gap};

  \node[font=\normalsize, anchor=north] at (3.0, -1.00) {(b) ReAct / GPT-OSS-120B};

\end{tikzpicture}

\caption{Consistency gap: normalized consistency (\Pk{5}/Mean@5) with and without consistency guidelines, by task difficulty.}
\label{fig:closing-gap}
\end{figure*}

\subsection{Generalization to Similar Tasks}
\label{sec:eval-similar-task}
A more demanding test is whether guidelines extracted from one task improve consistency on a different but similar task. Each AppWorld scenario consists of three similar task variants $v1,v2,v3$ that differ in parameter values and other details; we derive guidelines from $v_1$ and inject them into runs of $v_2$ and $v_3$ so each of the 168 tasks is evaluated twice, against guidelines from each of its scenario's other two variants (336 evaluations total)

\paragraph{GPT-4.1.}
Sibling-task guidelines lift aggregate \Pk{5} by $+13.0$ pp (Table~\ref{tab:results-gpt41}), only $3$ pp below same-task; Hard shows a small Mean@5 regression ($-0.8$ pp), likely because some sibling guidelines target API patterns that differ subtly in the sibling variant. That transfer occurs at all suggests the analyzer surfaces recurring patterns (e.g., paginated retrieval, credential management) rather than task-idiosyncratic ones.

\paragraph{GPT-OSS-120B.}
Sibling-task guidelines lift aggregate \Pk{5} by $+8.7$ pp (Table~\ref{tab:results-oss}), \emph{exceeding} the $+6.0$ pp same-task gain, with all four difficulty levels improving and no regressions; Hard tasks move from $0.00$ to $0.29$ normalized consistency, the largest single-cell gain across both tables.

\paragraph{Consistency Gap.}
Figure~\ref{fig:closing-gap}'s dark teal bars show similar-task gap reduction tracking same-task closely for GPT-4.1 (aggregate gap $0.17$ vs.\ $0.15$), and substantially outperforming same-task on GPT-OSS-120B's Hard tasks ($0.29$ vs.\ $0.13$ closure from zero). Overall, guidelines reliably push normalized consistency upward across difficulty levels, with similar-task transfer matching or exceeding same-task closure in most cases.

\subsection{Trajectory Consistency as an Outcome Predictor}
\label{sec:analyzer-validity-results}

Section~\ref{sec:traj-aggregation} noted that aggregate trajectory consistency correlates with grader outcome; we quantify this using AUROC of aggregate consistency as a binary classifier of trajectory pass/fail. On 50 AppWorld \emph{test\_challenge} tasks with a single ReAct/GPT-4.1 trajectory each, mean aggregate consistency reaches AUROC $0.699$, an acceptable predictor. This makes aggregate consistency a useful signal for the Guideline Generator: rather than relying solely on the LLM's own unaided judgment of whether a trajectory succeeded or failed, the generator can draw on this empirical stability measure to support that judgment, which is otherwise unavailable in production settings that lack a ground-truth outcome rating.

\section{Related Work}
\label{sec:relatedwork}

\textbf{Self-consistency, inference scaling, and uncertainty quantification.}
Self-consistency~\cite{wang2023selfconsistency} samples multiple chains of thought and selects the majority answer; inference-scaling work~\cite{snell2024scaling} and reasoning models such as OpenAI's o1 family generalize the same intuition, and agent variants use tests or an LLM judge to select among rolled-out solutions. We differ on two axes: resampling happens \emph{offline} rather than at runtime, and the objective is consistency, not accuracy - variability is a diagnostic signal for where the prompt should be hardened, not a voting mechanism. Predictive entropy~\cite{malinin2021uncertainty} overstates uncertainty when responses vary in wording but agree in meaning, since it measures spread over surface forms rather than over meaning; semantic entropy~\cite{kuhn2023semantic, farquhar2024detecting} addresses this via clustering; our decomposed structured-response score (Section~\ref{sec:step-consistency}) targets the partially-structured nature of agent responses similarly.

\textbf{Agentic memory.}
Recent surveys~\cite{zhang2024survey, du2025rethinking} chart this growing space. \textsc{Mem0}~\cite{chhikara2025mem0}, A-Mem~\cite{xu2025amem}, and MemGPT~\cite{packer2023memgpt} store and retrieve conversational facts; Agent Workflow Memory~\cite{wang2024awm}, AgentRR~\cite{feng2025agentrr}, MemP~\cite{fang2025memp}, ReasoningBank~\cite{cai2025reasoningbank}, ACE~\cite{zhang2025ace}, and Memento~\cite{zhou2025memento} extract reusable workflows from traces, though naive accumulation can propagate errors~\cite{xiong2025memory}. We build on \citet{fang2026trajectory}, adding a consistency-targeted extraction signal atop its storage and retrieval machinery.

\textbf{Reliability, reproducibility, and observability.}
AppWorld~\cite{trivedi2024appworld} is one of several benchmarks reporting multi-run statistics, and Ye et al.~\cite{ye2026fragility} show that such run-to-run variance is itself an underspecified property of self-improving, memory-based agents, sensitive to task order and evaluation protocol; our guideline-injection loop targets exactly this source of fragility. Our \Pk{k} vs.\ Mean@$k$ formulation generalizes pass@$k$ from code generation~\cite{chen2021codex} to per-difficulty granularity, and lands within the broader reliability taxonomy of Rabanser et al.~\cite{rabanser2026science}, whose twelve metrics span consistency, robustness, predictability, and safety; our gap is a targeted probe of their consistency axis. CUGA~\cite{marreed2025cuga} adopts a related notion through policy memory and can consume our guidelines directly. Closest in spirit is concurrent work by Jha et al.~\cite{jha2026reliability}, who identify a \emph{reliability gap} between Pass@$k$ and Majority@$k$ addressed via a graph-guided architecture; we target the same variability at the prompt level. Our analyzer also functions as an observability tool, complementary to work on runtime logs~\cite{moshkovich2025observability}, process discovery~\cite{fournier2025observability}, process supervision~\cite{liu2026trajad}, and interactive debugging~\cite{hutter2026agentstepper}, but with a focus on surfacing flip-prone points for targeted guideline generation rather than runtime intervention.

\section{Discussion and Limitations}
\label{sec:discussion}

\textbf{Guideline validation.}
The pipeline described above admits every generated guideline directly into memory. A natural extension, which we leave to future work, is a validation stage between generation and memory admission: before a guideline is stored, re-run the analyzer's resampling procedure at its source step with the guideline injected into the prompt, and admit it only if step consistency measurably improves and does not regress on a held-out set of related trajectories. This would guard against guidelines that are vague, conflict with existing instructions, or overfit to incidental details of the trajectory they were derived from, a known risk in agentic memory more broadly~\cite{xiong2025memory}. 

\textbf{Generality and future directions.}
We evaluated a ReAct agent with GPT-4.1 and GPT-OSS-120B; the framework is agent-agnostic by design.  A cross-stack study, including models from other model families and agents with existing policy memory such as CUGA~\cite{marreed2025cuga}, is our next step.

\textbf{Limitations.}
Our \Pk{5} comparisons assume comparable platform-side non-determinism between baseline and guideline-injected runs, which we cannot directly measure or control. We do not claim the framework eliminates inconsistency (a substantial gap remains, especially on Hard tasks) or that the analyzer is a verifier: it scores \emph{stability}, not correctness, and a confidently-wrong decision looks consistent to it. Our default resampling budget ($N=30$ per inference step) multiplies token cost roughly $30\times$ per trajectory; reducing this cost without degrading guideline quality, e.g., via adaptive resampling that concentrates samples on candidate flat steps, is an important direction for future work.

\section{Conclusion}
\label{sec:conclusion}

Current LLM agents pass tasks ``on average'' in a way that masks how often they fail to pass the \emph{same} task on repeat. We presented a framework that reduces this consistency gap through consistency-targeted episodic memory guidelines: a black-box Consistency Analyzer resamples each inference point to surface decisions at risk of flipping, and a Guideline Generator converts them into consistency guidelines. On AppWorld with ReAct and GPT-4.1, the pipeline lifts \Pk{5} by 16 points same-task and 13 points similar-task with no accuracy degradation; aggregate trajectory consistency alone correlates with trajectory pass/fail (AUROC $\approx 0.69$), letting the analyzer support the generator's own judgment of trajectory outcome where no ground-truth rating exists.

\appendix

\begin{table*}[t]
\centering
{\Large\bf Appendix\par}
\vspace{6pt}
\refstepcounter{section}\label{app:guideline-examples}
{\Large\bf \thesection\quad Example Consistency Guidelines\par}
\vspace{4pt}
\begin{minipage}{\textwidth}
\small\noindent Section~\ref{sec:guidelines} references the example guidelines below, generated by GPT-4.1 from the consistency scorecard for AppWorld task \texttt{0a9d82a\_2} (``What is my longest limited-screen-time-to-1-hr habit streak, in number of days, as per my Simple Note habit tracking logs?'').
\end{minipage}
\vspace{6pt}
\begin{lstlisting}[basicstyle=\footnotesize\ttfamily]
[strategy] Always verify API parameter requirements and expected input/output formats by consulting the API documentation before making calls, especially for search and data retrieval APIs.

[strategy] When retrieving paginated data, implement a robust loop that continues fetching pages until all results are retrieved, checking for pagination end conditions as specified in the API documentation.

[recovery] Explicitly handle authentication and token management by retrieving fresh credentials as needed and verifying token validity before making authenticated API calls.

[optimization] After retrieving and parsing data, implement validation checks before performing calculations or returning results.
\end{lstlisting}
\end{table*}

\begin{table*}[t]
\centering
\refstepcounter{section}\label{app:results}
{\Large\bf \thesection\quad Full Per-Difficulty Results\par}
\vspace{4pt}
\begin{minipage}{\textwidth}
\small\noindent Tables~\ref{tab:results-gpt41} and~\ref{tab:results-oss} report the complete per-difficulty breakdown (baseline, same-task guidelines, and similar-task guidelines) underlying the aggregate numbers discussed in Section~\ref{sec:evaluation}.
\end{minipage}
\vspace{6pt}
\caption{Consistency results on AppWorld \texttt{test\_normal}, ReAct / GPT-4.1.}
\label{tab:results-gpt41}
\small
\setlength{\tabcolsep}{5pt}
\renewcommand{\arraystretch}{1.15}
\begin{tabular}{lccccccccc}
\toprule
& \multicolumn{3}{c}{Baseline}
& \multicolumn{3}{c}{$+$ Guidelines: Same Task}
& \multicolumn{3}{c}{$+$ Guidelines: Similar Task} \\
\cmidrule(lr){2-4}\cmidrule(lr){5-7}\cmidrule(lr){8-10}
Difficulty & \Pk{5} & Mean@5 & Consistency
           & \Pk{5} & Mean@5 & Consistency
           & \Pk{5} & Mean@5 & Consistency \\
\midrule
Aggregate & 53.0\% & 77.4\% & 0.68
  & \textbf{69.0\%} & \textbf{81.0\%} & \textbf{0.85}
  & \textbf{66.0\%} & \textbf{79.5\%} & \textbf{0.83} \\
$\Delta$ & & & & \textit{+16.0 pp} & \textit{+3.6 pp} & \textit{+0.17}
               & \textit{+13.0 pp} & \textit{+2.1 pp} & \textit{+0.15} \\
\cmidrule(lr){1-10}
Easy      & 77.2\% & 94.7\% & 0.82
  & \textbf{89.4\%} & \textbf{98.2\%} & \textbf{0.91}
  & \textbf{90.3\%} & \textbf{97.3\%} & \textbf{0.93} \\
$\Delta$ & & & & \textit{+12.2 pp} & \textit{+3.5 pp} & \textit{+0.10}
               & \textit{+13.1 pp} & \textit{+2.6 pp} & \textit{+0.11} \\
\cmidrule(lr){1-10}
Medium    & 52.1\% & 77.1\% & 0.68
  & \textbf{75.0\%} & \textbf{83.3\%} & \textbf{0.90}
  & \textbf{66.6\%} & \textbf{83.3\%} & \textbf{0.80} \\
$\Delta$ & & & & \textit{+22.9 pp} & \textit{+6.2 pp} & \textit{+0.22}
               & \textit{+14.5 pp} & \textit{+6.2 pp} & \textit{+0.12} \\
\cmidrule(lr){1-10}
Hard      & 31.7\% & 61.9\% & 0.51
  & \textbf{46.0\%} & \textbf{63.4\%} & \textbf{0.73}
  & \textbf{43.6\%} & 61.1\%          & \textbf{0.71} \\
$\Delta$ & & & & \textit{+14.3 pp} & \textit{+1.6 pp}   & \textit{+0.21}
               & \textit{+11.9 pp} & \textit{$-$0.8 pp} & \textit{+0.20} \\
\bottomrule
\end{tabular}
\end{table*}

\begin{table*}[t]
\centering
\caption{Consistency results on AppWorld \texttt{test\_normal}, ReAct / GPT-OSS-120B.}
\label{tab:results-oss}
\small
\setlength{\tabcolsep}{5pt}
\renewcommand{\arraystretch}{1.15}
\begin{tabular}{lccccccccc}
\toprule
& \multicolumn{3}{c}{Baseline}
& \multicolumn{3}{c}{$+$ Guidelines: Same Task}
& \multicolumn{3}{c}{$+$ Guidelines: Similar Task} \\
\cmidrule(lr){2-4}\cmidrule(lr){5-7}\cmidrule(lr){8-10}
Difficulty & \Pk{5} & Mean@5 & Consistency
           & \Pk{5} & Mean@5 & Consistency
           & \Pk{5} & Mean@5 & Consistency \\
\midrule
Aggregate & 10.1\% & 33.9\% & 0.30
  & \textbf{16.1\%} & \textbf{38.7\%} & \textbf{0.42}
  & \textbf{18.8\%} & \textbf{40.5\%} & \textbf{0.46} \\
$\Delta$ & & & & \textit{+6.0 pp}  & \textit{+4.8 pp} & \textit{+0.12}
               & \textit{+8.7 pp}  & \textit{+6.6 pp} & \textit{+0.17} \\
\cmidrule(lr){1-10}
Easy      & 28.1\% & 66.7\% & 0.42
  & \textbf{36.8\%} & \textbf{75.4\%} & \textbf{0.49}
  & \textbf{43.0\%} & \textbf{74.6\%} & \textbf{0.58} \\
$\Delta$ & & & & \textit{+8.7 pp}  & \textit{+8.7 pp} & \textit{+0.07}
               & \textit{+14.9 pp} & \textit{+7.9 pp} & \textit{+0.16} \\
\cmidrule(lr){1-10}
Medium    &  2.1\% & 27.1\% & 0.08
  & \textbf{10.4\%} & \textbf{29.2\%} & \textbf{0.36}
  & \textbf{8.3\%}  & \textbf{31.2\%} & \textbf{0.27} \\
$\Delta$ & & & & \textit{+8.3 pp}  & \textit{+2.1 pp} & \textit{+0.28}
               & \textit{+6.2 pp}  & \textit{+4.1 pp} & \textit{+0.19} \\
\cmidrule(lr){1-10}
Hard      &  0.0\% &  9.5\% & 0.00
  & \textbf{1.6\%}  & \textbf{12.7\%} & \textbf{0.13}
  & \textbf{4.8\%}  & \textbf{16.7\%} & \textbf{0.29} \\
$\Delta$ & & & & \textit{+1.6 pp}  & \textit{+3.2 pp} & \textit{+0.13}
               & \textit{+4.8 pp}  & \textit{+7.2 pp} & \textit{+0.29} \\
\bottomrule
\end{tabular}
\end{table*}

\clearpage
\interlinepenalty=10000
\brokenpenalty=10000
\raggedbottom
\enlargethispage{40\baselineskip}
\bibliography{consistency}

\end{document}